\def\year{2019}\relax
\documentclass[letterpaper]{article} %DO NOT CHANGE THIS
\usepackage{aaai19}  %Required
\nocopyright 
\usepackage{times}  %Required
\usepackage{helvet}  %Required
\usepackage{courier}  %Required
\usepackage{url}  %Required
\usepackage{graphicx}  %Required
\usepackage{subfig}
\usepackage{enumitem}
\usepackage{amsmath}
\usepackage{multirow}
\usepackage[linesnumbered,ruled,vlined]{algorithm2e}
\newcommand*\system{\textsc{Rona}}
\newtheorem{thm}{Theorem}
\newtheorem{definition}{Definition}
\SetKwInOut{Parameter}{Parameter}

\usepackage[table,xcdraw]{xcolor}
\usepackage{mathtools}
\usepackage{amsfonts}
\usepackage{balance}
\usepackage{flushend}

\begin{document}
\title{Private Model Compression via Knowledge Distillation}
\author{Ji Wang\textsuperscript{1}, Weidong Bao\textsuperscript{1}, Lichao Sun\textsuperscript{2}, Xiaomin Zhu\textsuperscript{1}\textsuperscript{3}, Bokai Cao\textsuperscript{4}, Philip S. Yu\textsuperscript{2}\textsuperscript{5}\\
\textsuperscript{1}{College of Systems Engineering, National University of Defense Technology, Changsha, China}\\
\textsuperscript{2}{Department of Computer Science, University of Illinois at Chicago, Chicago, USA}\\
\textsuperscript{3}{State Key Laboratory of High Performance Computing, National University of Defense Technology, Changsha, China}\\
\textsuperscript{4}{Facebook Inc., Menlo Park, USA}\\
\textsuperscript{5}{Institute for Data Science, Tsinghua University, Beijing, China}\\
\{wangji,wdbao,xmzhu\}@nudt.edu.cn,
lsun29@uic.edu,
caobokai@fb.com,
psyu@uic.edu
}
\maketitle
\begin{abstract}
The soaring demand for intelligent mobile applications calls for deploying powerful deep neural networks (DNNs) on mobile devices. However, the outstanding performance of DNNs notoriously relies on increasingly complex models, which in turn is associated with an increase in computational expense far surpassing mobile devices' capacity. What is worse, app service providers need to collect and utilize a large volume of users' data, which contain sensitive information, to build the sophisticated DNN models. Directly deploying these models on public mobile devices presents prohibitive privacy risk. To benefit from the on-device deep learning without the capacity and privacy concerns, we design a private model compression framework \system. Following the knowledge distillation paradigm, we jointly use hint learning, distillation learning, and self learning to train a compact and fast neural network. The knowledge distilled from the cumbersome model is adaptively bounded and carefully perturbed to enforce differential privacy. We further propose an elegant query sample selection method to reduce the number of queries and control the privacy loss. A series of empirical evaluations as well as the implementation on an Android mobile device show that \system\ can not only compress cumbersome models efficiently but also provide a strong privacy guarantee. For example, on SVHN, when a meaningful $(9.83,10^{-6})$-differential privacy is guaranteed, the compact model trained by \system\ can obtain 20$\times$ compression ratio and 19$\times$ speed-up with merely 0.97\% accuracy loss.
\end{abstract}

\section{Introduction}
Recent years have witnessed impressive breakthroughs of deep learning in various areas. Thanks to the large volume of data and the easy availability of computational resources, customized deep learning models attain unprecedented performance that beats many records of traditional machine learning algorithms.

The significant progress in deep learning, on the other hand, is notorious for its dependence on cumbersome models which require massive data to learn their millions of parameters. This major drawback restricts the large-scale deployment of deep learning applications, especially the deployability of deep neural network on mobile devices such as smartphones, wearable devices, and medical monitors \cite{cao2017deepmood,li2016droidclassifier}. Mobile application service providers are facing a series of challenges to widely adopt DNNs in their mobile apps.

\textbf{Capacity bottleneck}. In spite of the great advances of mobile chips and mobile batteries, the limited-capacity nature of mobile devices still imposes the intrinsic bottleneck making resource-demanding applications remain off bounds \cite{wang2018not}. The modern DNNs with millions of parameters often require prohibitive runtime to execute on computationally limited devices. What is worse, the massive floating point operations during the execution aggravate the burden of processing chips and easily dominate the whole system energy consumption, which widens the chasm between DNNs and their deployability on mobile devices. It is difficult to directly adopt powerful DNNs on mobile systems so far \cite{lee2017technology}.

\textbf{Data privacy and intellectual piracy concerns}. Developing a strong predictive DNN needs abundant data. The more data accessible, the more effective and powerful a DNN will be. In practice, app service providers usually collect and utilize a large volume of data from users, which often contains sensitive information, to build their sophisticated DNN models. Directly releasing these models trained by users' data presents potential privacy issues because the adversary can recover sensitive information encoded in the public models \cite{hitaj2017deep,abadi2016deep}. It is illegal to share individuals' data or models directly in many domains like medicine \cite{hipaa2013}. Apart from data privacy, releasing DNN models may invade app service providers' right due to intellectual piracy. It is sometimes not appealing for app service providers to share their valuable and highly tuned models \cite{osia2017a}.

In order to deploy efficient DNNs on mobile devices, academia and industry put forward a number of model compression methods among which knowledge distillation plays a key role \cite{bucila2006model}. In knowledge distillation, the knowledge embedded in the cumbersome model, known as the teacher model, is distilled to guide the training of a smaller model called the student model. The student model has a different architecture and fewer parameters, but can achieve comparable performance by mimicking the behavior of the cumbersome model. Other compression methods like quantization and low-rank factorization \cite{han2015deep,howard2017mobile} are complementary to knowledge distillation and can also be used to further reduce the size of student models, which is beyond the scope of this paper. Despite the encouraging compression rate, the privacy concern has not been fully resolved by the current knowledge distillation methods yet. 

In this paper, we introduce the rigorous standard of differential privacy \cite{dwork2011diff} and design a p\textbf{R}ivate m\textbf{O}del compressio\textbf{N} fr\textbf{A}mework, named \system, to overcome the aforementioned challenges and promote the adoption of deep learning in mobile applications.

Both model compression and data privacy are considered in the proposed framework. We assume that the app service provider has trained a powerful cumbersome model based on the sensitive data and the public data. Following the knowledge distillation paradigm, \system\ uses only the public data to train the small student model whose feature representations are encouraged to be similar to those of the cumbersome model. Then the small student model is released to mobile application users while the cumbersome model as well as the sensitive data are retained by the app service provider. Intuitively, the privacy is preserved because the training of the student model does not depend on the sensitive data; neither the cumbersome model nor the sensitive data is exposed to the public or accessible to the adversary. Mere intuition, however, is not sufficient. To provide provable privacy guarantee, we carefully perturb the knowledge distilled from the cumbersome model to satisfy the standard of differential privacy.

It is a nontrivial problem to jointly compress large DNNs, preserve privacy, and control model performance loss. A set of novel methods are presented to solve this problem. Our main contributions are four-fold:

\begin{itemize}[leftmargin=*,noitemsep,topsep=0pt]
\item \textbf{A framework promoting deep learning in mobile applications}. We take the key constraints of adopting DNNs in mobile applications, \textit{i.e.}, privacy, performance, and overhead, into consideration, and design a privacy-preserving framework for training compact neural networks via knowledge distillation\footnote{This paper uses the multi-class image recognition as the application scenario due to its wide usage in mobile apps.}. To the best of our knowledge, the proposed \system\ is the first framework that applies the knowledge distillation to model compression with meaningful privacy guarantee.
\item \textbf{A differentially private knowledge distillation}. To theoretically guarantee privacy, we propose a new mechanism to perturb the knowledge distillation from the sense of differential privacy. Different from the existing sample-by-sample query mode, our proposed mechanism makes queries in batch mode to reduce the number of queries. The batch loss responded by the teacher model is clipped by the adaptive norm bound and then carefully perturbed to preserve privacy.
\item \textbf{A query sample selection method for knowledge distillation}. The privacy loss depends on the number of queries. To control the student model's access to the teacher model, the number of samples used during knowledge distillation should be reduced. Hence, we present a query sample selection method to select a subset of samples such that the knowledge distilled over the subset is competitive over the whole samples.
\item \textbf{Thorough empirical evaluation}. We evaluate the proposed \system\ by using three standard benchmarks that are widely used in knowledge distillation works. The results demonstrate the effectiveness of the above novel methods, bringing significant improvement in training small models with rigorous privacy guarantee. Our code is open-sourced at \url{https://github.com/jwanglearn/Private_Compress}.
\end{itemize}

\section{Preliminary and Related Work}
\textbf{Deep neural network compression}. To squeeze DNNs into mobile devices, DNN compression attracts intense attention. The compression methods can be broadly classified into four categories: parameter sharing, network pruning, low-rank factorization, and knowledge distillation \cite{han2015deep,howard2017mobile,cheng2018learning}. The former three methods mainly attempt to reduce the size of a given model, without significantly changing the architecture of the model. The last one, knowledge distillation, uses the knowledge captured by the cumbersome model to guide the training of a smaller model in a teacher-student paradigm \cite{bucila2006model}. Hinton et al. \cite{hinton2015distilling} used the class probabilities generated by the teacher as a soft target to train the student model. Romero et al. \cite{romero2015fitnets} extended this work by training the student to mimic the teacher's intermediate representation. Based on these two works, a framework for compressing object detection models was designed by Chen et al. \cite{cheng2018learning}, trying to solve the class imbalance problem. Few existing model compression methods took the privacy issue into consideration. The teacher model can be queried as many times as necessary during training, which is infeasible if we want to preserve privacy.

\textbf{Differentially private deep learning}. Due to the critical need of respecting privacy, privacy-preserving data analysis has become an emerging topic of interests. One state-of-the-art privacy standard is differential privacy \cite{dwork2011diff} which provides provable privacy guarantee. 

\begin{definition}~\cite{dwork2011diff}
  A randomized mechanism $\mathcal{M}$ is $(\epsilon,\delta)$-differentially private, iff for any adjacent input $d$ and $d'$, and any output $S$ of $\mathcal{M}$,
  \begin{equation}
    \Pr [\mathcal{M} (d) = S] \le {e^\epsilon } \cdot \Pr [\mathcal{M} (d') = S]+\delta. 
  \end{equation}
\end{definition}

Typically, $d$ and $d'$ are adjacent inputs when they are identical except for only one data item. The parameter $\epsilon$ denotes the privacy budget \cite{dwork2011diff}, controlling the privacy loss of $\mathcal{M}$. A smaller value of $\epsilon$ enforces a stronger privacy guarantee.

%=====================================================
\begin{figure*}[tb]
\centering
\includegraphics[width=5.4in]{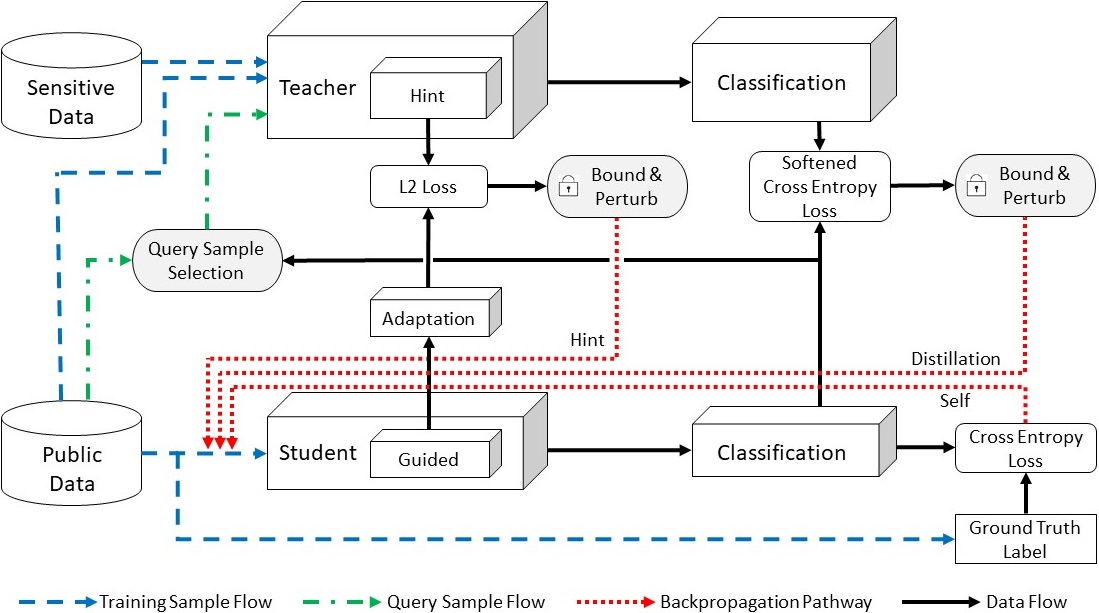}
\caption{The overview of \system. \system\ is implemented by the mobile app service providers. Only the student model is released to the public while the sensitive data and the teacher model are retained by the mobile app service provider.} \label{fig:framework}
\end{figure*}
%======================================================

For a deterministic function $f$, the $(\epsilon,\delta)$-differential privacy is generally enforced by injecting random noise calibrated to the $f$'s sensitivity $\Delta f$, $\Delta f = \max_{d,d'}\left\| {f(d) - f(d')} \right\|$. For example, the Gaussian mechanism is given by,

\begin{thm}\label{thm:gaussian}~\cite{dwork2014the}
Suppose function $f$ with L2 norm sensitivity $\Delta _2 f$, a randomized mechanism $\mathcal{M} _f(d)$:
\begin{equation}
\mathcal{M}_f (d) = f(d)+\mathcal{N}(0, {\Delta _2 f}^2\sigma^2),
\end{equation}
where $\mathcal{N}(0, {\Delta _2 f}^2\sigma^2)$ is a random variable sampled from the Gaussian distribution with mean 0 and standard deviation ${\Delta _2 f}\sigma$. $\mathcal{M}_f(d)$ is $(\epsilon, \delta)$-differentially private if $\sigma  \ge \sqrt {2\ln (1.25/\delta )} /\epsilon $ and $\epsilon < 1$.
\end{thm}

Differential privacy provides guaranteed privacy which cannot be compromised by any algorithm \cite{dwork2011diff}. It is increasingly adopted as the standard notion of privacy \cite{beimel2014bounds}. Abadi et al. \cite{abadi2016deep} presented a differentially private SGD and designed a new technique to track the privacy loss. Papernot et al. \cite{papernot2017semi,papernot2018scalable} proposed a general framework for private knowledge transfer in the sample-by-sample mode where the student was trained to predict an output generated by noisy voting among the teachers. This framework only used the output label to train the student and cannot be used to compress DNNs efficiently. Apart from private training, Triastcyn et al. \cite{triastcyn2018generating} added a Gaussian noise layer in the discriminator of GAN to generate differentially private data. Recently, Wang et al. \cite{wang2018not} designed a private inference framework across mobile devices and cloud servers to free mobile devices from complex inference tasks. However, this framework heavily depended on the network accessibility. Few current works applied differential privacy and model compression to enabling on-device deep learning as well as preserving privacy.

\section{The Proposed Framework}
The framework \system\ is presented in this section. We first give the overview of \system. Then we detail three key modules: (1) the model compression based on knowledge distillation, (2) the differentially private knowledge perturbation, and (3) the query sample selection.

\subsection{Overall Structure}
The overview of \system\ is given in Fig. \ref{fig:framework}. To better capture the knowledge embedded in the cumbersome teacher model, we jointly use the hint learning \cite{romero2015fitnets}, the distillation learning \cite{hinton2015distilling}, and the self learning to train a small student model. Meanwhile, both the hint loss and the distillation loss are carefully bounded and perturbed by random noises that are consistent with differential privacy. Since more queries to the teacher model incur higher privacy loss, an elegant query sample selection method is designed in \system\ to select a subset of samples from the entire public samples.

The sensitive data is only used to train the complex teacher model which is not released to the public. It is obvious that the sensitive data are insulated from the explicit invasion of privacy as the student model has no access to it. Further, the information generated by the teacher model, \textit{i.e.}, the hint loss and the distillation loss, are perturbed by additional noises. All the information relating to the sensitive data and the teacher model are isolated or well protected before releasing.

\subsection{Model Compression}
In order to learn a compact and fast DNN model that can be adopted in mobile applications, we propose a stage-wise knowledge distillation method. This method enables the small student model to capture not only the information in ground truth labels but also the information distilled from the cumbersome teacher model.

Hint learning stage: We start by teaching the student model how to extract features from the input data. The intermediate representation of the teacher model is used as a hint to guide the training of the student model. Analogously, a hidden layer of the student model is chosen as the guided layer that learns from the teacher's guidance. We train the student model from the first hidden layer up to the guided layer by minimizing the L2 loss function:
\begin{equation}\label{eqn:hint-loss}
\mathcal{ L }_{ hint }({ \textbf{x} }_{ q },\textbf{z}_{ h };{ \Theta  }_{ g },{ \Theta  }_{ a })=\frac { 1 }{ 2 } { \left\| r(g({ \textbf{x} }_{ q };{ \Theta  }_{ g });{ \Theta  }_{ a })-\textbf{z}_{ h } \right\|  }^{ 2 },
\end{equation}
where $g(\cdot;\Theta_g)$ represents the student model up to the guided layer with parameter $\Theta_g$, $\textbf{x}_{ q }$ denotes the query samples, and $\textbf{z}_h$ is the output of the teacher's hint layer over the query samples. As the scale of the guided layer is usually different from that of the hint layer, we introduce an adaptation layer on the top of the guided layer to match the scale of the guided and hint layers. The adaptation layer is represented by $r(\cdot;\Theta_a)$ with parameter $\Theta_a$ learned during the hint learning. If both guided and hint layers are fully connected layers, we add a fully connected layer as the adaptation layer. If both guided and hint layers are convolutional, we add $1\times 1$ convolutions instead to reduce the number of parameters.

Hint learning is introduced to teach the student how to extract general features. It makes the student model lack flexibility to choose a higher hidden layer as the guided layer. So, we select the student's middle layer as the guided layer in our case.

%=====================================================
\begin{algorithm}[tb]
\SetAlgoVlined
\small
% \SetKw{Parameter}{Parameters:}
\Parameter{Hint learning epochs $T_{h}$; Distillation learning epochs $T_{d}$; Self learning epochs $T_{s}$; Iterations \textit{R}; Hint and distillation learning batch size $S$; Self learning batch size $S'$.}
\For{$t\gets0$ \KwTo $T_h-1$}{
    $\textbf{x}_q\gets$ \texttt{random\_select}($\textbf{x}_p$)\;
    \For{$i\gets0$ \KwTo $|\textbf{x}_q|/S$}{
        Sample a batch $\textbf{x}^{(i)}_q$ of size $S$ from $\textbf{x}_q$\;
        Compute hint loss $\mathcal{L}^{(i)}_{hint}$ over $\textbf{x}^{(i)}_q$ by Eq. (\ref{eqn:hint-loss})\;
        $\widetilde{\mathcal{L}}^{(i)}_{hint}\gets$ \texttt{privacy\_sanitize}($\mathcal{L}^{(i)}_{hint}$)\;
        Backpropagate by $\widetilde{\mathcal{L}}^{(i)}_{hint}$ to update $\Theta_g$ and $\Theta_a$\;
    }
}
\For{$r\gets0$ \KwTo $R-1$}{
    \For{$t\gets0$ \KwTo $T_s-1$}{
    \For{$i\gets0$ \KwTo $|\textbf{x}_p|/S'$}{
        Sample a batch $\textbf{x}^{(i)}_p$ of size $S'$ from $\textbf{x}_p$\;
        Compute self loss $\mathcal{L}^{(i)}_{self}$ over $\textbf{x}^{(i)}_p$ by Eq. (\ref{eqn:self-loss})\;
        Backpropagate by $\mathcal{L}^{(i)}_{self}$ to update $\Theta_s$\;
    }
    }
    \For{$t\gets0$ \KwTo $T_d-1$}{
    $\textbf{x}_q\gets$ \texttt{query\_select}($\textbf{x}_p$, $\Theta_s$)\;
    \For{$i\gets0$ \KwTo $|\textbf{x}_q|/S$}{
        Sample a batch $\textbf{x}^{(i)}_q$ of size $S$ from $\textbf{x}_q$\;
        Compute distill loss $\mathcal{L}^{(i)}_{distill}$ over $\textbf{x}^{(i)}_q$ by Eq. (\ref{eqn:distill-loss})\;
        $\widetilde{\mathcal{L}}^{(i)}_{distill}\gets$ \texttt{privacy\_sanitize}($\mathcal{L}^{(i)}_{distill}$)\;
        Backpropagate by $\widetilde{\mathcal{L}}^{(i)}_{distill}$ to update $\Theta_s$\;
    }
    }
}
\small\caption{Compact Student Model Training}
\end{algorithm}
%======================================================

Distillation and self learning stage: We train the whole student model by using the distillation and self learning in this stage. Let $\textbf{z}_t$ be the output of the teacher's final hidden layer, also called logits, over the query samples, we use the soften probability \cite{hinton2015distilling} $\textbf{P}^\tau_t$ as the knowledge: $\textbf{P}^\tau_t=softmax(\textbf{z}_t/\tau)$, where $\tau$ is the temperature parameter that is normally set as 1. A higher $\tau$ can make the teacher model generate soften probabilities such that the classes whose normal probabilities are near zero will not be ignored. The soften probability contains the knowledge about the latent relationship between different classes. The student model is trained to learn this knowledge by minimizing the distillation loss over the query samples:
\begin{equation}\label{eqn:distill-loss}
\mathcal{ L }_{ distill }({ \textbf{x} }_{ q },\textbf{P}^\tau_{ t };{ \Theta  }_{ s })=\mathcal{H}(\textbf{P}^\tau_s, \textbf{P}^\tau_{ t };{ \Theta  }_{ s }),
\end{equation}
where $\mathcal{H}$ is the cross-entropy, and $\Theta_s$ is the parameters of the student model. Here, $\textbf{P}^\tau_s$ is the student's soften probability over the query samples $\textbf{x} _{ q }$, $\textbf{P}^\tau_s = softmax(\textbf{z}_s/\tau)$, where $\textbf{z}_s$ is the student's logits.

Different from the classical use of knowledge distillation, we introduce the self learning process where the ground truth labels of all public samples are used to train the student model by minimizing the self loss:
\begin{equation}\label{eqn:self-loss}
\mathcal{ L }_{ self }({ \textbf{x} _{ p }, \textbf{y} }_{ p };{ \Theta  }_{ s })=\mathcal{H}(\textbf{P}_s, \textbf{y} _{ p };{ \Theta  }_{ s }),
\end{equation}
where $\textbf{x} _{ p }$ and $\textbf{y} _{ p }$ are the public samples and the ground truth labels, respectively, and $\textbf{P}_s$ is the normal probability generated by the student model with $\tau=1$.

For the privacy-preserving reason, the number of distillation-learning epochs should be controlled (detailed in the next section). Nonetheless, the self learning does not use any information relating to the sensitive data, and hence it does not aggregate the privacy loss. The number of self-learning epochs can be arbitrarily large. It is inappropriate to combine the distillation loss and the self loss together as a general loss for training. To jointly apply the two learning methods, we mimic the way how a real student learns from her teacher. The student model first learns by itself to minimize the self loss. Then, it selects some samples to query the teacher model, and learns from the teacher by minimizing the distillation loss. This procedure repeats until the convergence or exceeding the privacy budget. Our experimental results show that the self learning can not only accelerate the training of the student model but also allow the distillation learning to avoid local minima. 

Algorithm 1 presents the two-stage knowledge distillation. The student model queries the teacher model in a batch-by-batch mode. During the hint learning, the student model learns to extract general features. Hence, we select the query samples from the public samples randomly rather than using the proposed sample selection method.

\subsection{Privacy Protection}
To enforce theoretical privacy guarantee, we inject random perturbation into the information that is related with the sensitive data and is used by the student model, \textit{i.e.}, the hint loss and the distillation loss. Algorithm 2 outlines the privacy-preserving function \texttt{privacy\_sanitize}. For each batch loss $\mathcal{L}^{(i)}$, we first bound the batch loss by a threshold $B$, ensuring that the sensitivity is not larger than $B$; then we inject Gaussian noise into the bounded batch loss.

%=====================================================
\begin{algorithm}[tb]
\SetAlgoVlined
\small
\KwIn{Batch loss $\mathcal{L}^{(i)}$.}
\Parameter{Noise scale $\sigma$; Bound threshold $B$.}
$\bar{\mathcal{L}}^{(i)}\gets \mathcal{L}^{(i)}/\max { (1,\frac { \left\| { \mathcal{L}}^{(i)} \right\|_2}{B})}$\;
$\widetilde{\mathcal{L}}^{(i)}\gets \bar{\mathcal{L}}^{(i)} + \mathcal{N}(0,\sigma^2B^2\textbf{I})$\;
\KwOut{Sanitized batch loss $\widetilde{\mathcal{L}}^{(i)}$.}
\small\caption{Function Privacy Sanitize}
\end{algorithm}
%======================================================

It is hard to estimate the sensitivity of the batch loss over the sensitive data. Therefore, we clip the max value of $ \left\| \mathcal{L}^{(i)} \right\|_2 $ within a given bound as shown in Line 1 of Algorithm 2. The value of $ \left\| \mathcal{L}^{(i)} \right\| _2$ is preserved if $\left\| { \mathcal{L}}^{(i)} \right\|_2\leq B$, whereas it is scaled down to $B$ if $\left\| { \mathcal{L}}^{(i)} \right\|_2> B$. After clipping, the sensitivity of $\bar{\mathcal{L}}^{(i)}$ is $B$. 

An overly large $B$ will incur excessive noise, while too small $B$ will lead to over truncation of the batch loss, both causing low utility of the sanitized batch loss. To solve this problem, we propose to use an adaptive norm bound. Specially, we train an auxiliary teacher model based on the public data. We constantly monitor the auxiliary batch loss between the auxiliary teacher model and the student model during training and set the average value of the auxiliary batch loss as the norm bound. In this manner, the norm bound $B$ changes adaptively during training. The empirical study shows considerable performance improvement brought by such an adaptive norm bound. As the norm bound is independent with the sensitive data, no privacy budget would be consumed by clipping the batch loss.

Gaussian noise is then added into the bounded batch loss to preserve privacy. According to Theorem \ref{thm:gaussian}, this randomized mechanism enforces $(\epsilon,\delta)$-differential privacy per query if we set $\sigma$ as $\sqrt { 2\ln(1.25/\delta ) } /\epsilon $. During the training of the student model, the teacher model is queried $T=(T_h+RT_d)|\textbf{x}_q|/S$ times. Using moments accountant \cite{abadi2016deep}, we have the following theorem:

\begin{thm}\label{thm:moments}
Given $\epsilon<c_1T$, $\delta>0$, where $c_1$ is a constant. Algorithm 1 can achieve $(\epsilon,\delta)$-differential privacy by setting $\sigma$ as:
\begin{equation}
    \sigma\geq c_2\frac{\sqrt{T\ln(1/\delta)}}{\epsilon},
\end{equation}
where $c_2$ is a constant.
\end{thm}

Due to the page limitation, we omit the proof here. Theorem \ref{thm:moments} indicates that a larger value of $T$ incurs a larger privacy budget $\epsilon$, namely more privacy loss, when $\sigma$ is fixed. To provide stronger protection, the value of $T$ should be controlled. Therefore, instead of querying the teacher over all public samples, we select the subset $\textbf{x}_q$ as the query samples. Nonetheless, it is obvious that the downsampling of public samples has an adverse impact on knowledge distillation. To alleviate this impact, we design a novel query sample selection method to select the critical samples in the next section.

\subsection{Query Sample Selection}
The sample selection problem can be categorized as the active learning problem. Different from the traditional setting where the samples are chosen one-by-one to query, our proposed student-teacher query works in a batch mode. We attempt to select a set of query samples such that the distilled knowledge over the query samples and that over the whole public samples are as close as possible. Formally, we try to minimize the difference between the distillation loss over the query samples and that over the whole public samples:
\begin{equation}\label{eq:objection}
\mathop {\min }\limits_{{\textbf{x}_q}:{\textbf{x}_q} \subset {\textbf{x}_p}} \left| {{\mathcal{L}_{distill}}({\textbf{x}_p},\textbf{P}_t^\tau ) - {\mathcal{L}_{distill}}({\textbf{x}_q},\textbf{P}_t^\tau )} \right|.
\end{equation}

Because we have no prior knowledge of $\textbf{P}_t^\tau$, the above optimization objective is not tractable. Instead, we try to optimize the upper bound of this objective as given below.

\begin{thm}\label{thm:core}
Given the public samples $\textbf{x}_p$ and the query samples $\textbf{x}_q$. If $\textbf{x}_q$ is $\lambda$ cover of $\textbf{x}_p $, ${{L_{distill}}({\textbf{x}_p},\textbf{P}_t^\tau )}=A$, we have,
\begin{equation}\label{eq:upbound}
\begin{aligned}
& \quad \left| {{{\mathcal L}_{distill}}({{\textbf{x}}_p},{\textbf{P}}_t^\tau ) - {{\mathcal L}_{distill}}({{\textbf{x}}_q},{\textbf{P}}_t^\tau )} \right| \\
& \le {\mathcal O}(\lambda ) + {\mathcal O}(\sqrt {{1 \over {|{{\textbf{x}}_p}|}}} ) + {\mathcal O}(A).
\end{aligned}
\end{equation}
\end{thm}

The proof is omitted here due to the page limitation. In Theorem \ref{thm:core}, ``$\textbf{x}_q$ is $\lambda$ cover of $\textbf{x}_p $'' indicates that the whole $\textbf{x}_p$ can be covered by a group of spheres\footnote{Here we use the concepts in 3D space to make it easy-to-understand. The distance is defined in the feature space of data.} centered at each sample in $\textbf{x}_q$ with radius $\lambda$. In the right-hand side (RHS) of Eq. (\ref{eq:upbound}), ${\mathcal O}(\sqrt {{1 / {|{{\textbf{x}}_p}|}}} ) + {\mathcal O}(A)$ is independent with $\textbf{x}_q$. Hence, we can minimize $\lambda$ to control the RHS of Eq. (\ref{eq:upbound}). Now the optimization of (\ref{eq:objection}) is converted into: $\min _{\textbf{x}_q:\textbf{x}_q \subset \textbf{x}_p}\lambda$. This problem is equivalent to the minimax location problem \cite{korupolu2000analysis},
\begin{equation}
\mathop {\min }\limits_{{{\bf{x}}_q}:{{\bf{x}}_q} \subset {{\bf{x}}_p}} \mathop {\max }\limits_{{x_i} \in {{\bf{x}}_q}} \mathop {\min }\limits_{{x_j} \in {{\bf{x}}_q},{x_j} \ne {x_i}} \mathcal{D} ({x_i},{x_j}| \Theta_s), 
\end{equation}
where $\mathcal{D} ({x_i},{x_j}| \Theta_s)$ denotes the distance between two samples. The KL-divergence between the output probabilities is used as the distance. We use a 2-$OPT$ greedy algorithm to solve this problem \cite{korupolu2000analysis}. Algorithm 3 outlines the function for selecting query samples.

%=====================================================
\begin{algorithm}[tb]
\SetAlgoVlined
\small
\KwIn{Public samples $\textbf{x}_p$; Student parameters $\Theta_s$.}
\Parameter{Number of query samples $N_q$.}
Initialize $\textbf{x}_q$ with a sample randomly chosen from $\textbf{x}_p$\;
\For{$n\gets0$ \KwTo $N_q-2$}{
${x_{sel}}\gets \mathop {\arg \max }\limits_{{x_i} \in {{\bf{x}}_p} - {{\bf{x}}_q}} \mathop {\min }\limits_{{x_j} \in {{\bf{x}}_q}} \Lambda ({x_i},{x_j}|\Theta_s)$\;
${{\bf{x}}_q}\gets {{\bf{x}}_q} \cup \{ {x_{sel}}\}$\;
}
\KwOut{Query samples $\textbf{x}_q$.}
\small\caption{Function Query Select}
\end{algorithm}
%======================================================

%=====================================================
\begin{figure*}[tb]
\centering
\subfloat[Hint Learning Epochs]{\includegraphics[width=2.15in]{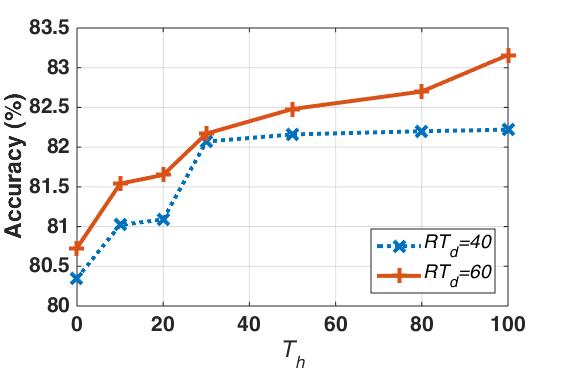}}\quad
\subfloat[Iterations]{\includegraphics[width=2.15in]{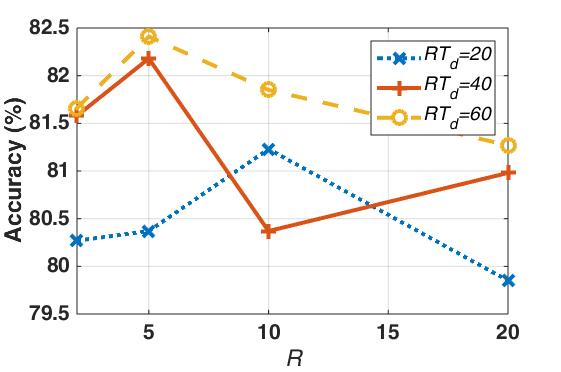}}\quad
\subfloat[Batch Size]{\includegraphics[width=2.15in]{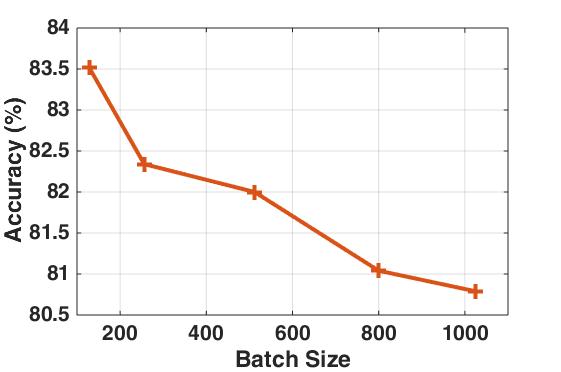}}\\
\subfloat[Noise Scale]{\includegraphics[width=2.15in]{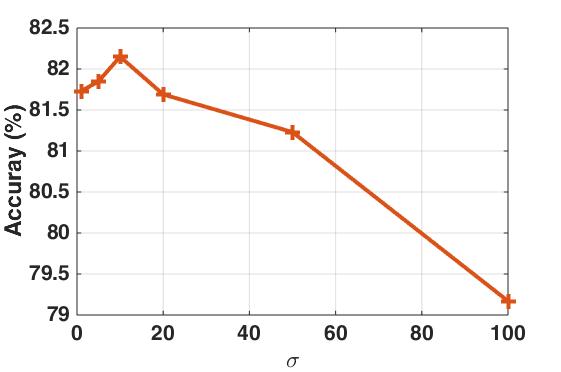}}\quad
\subfloat[\# of Query Samples]{\includegraphics[width=2.15in]{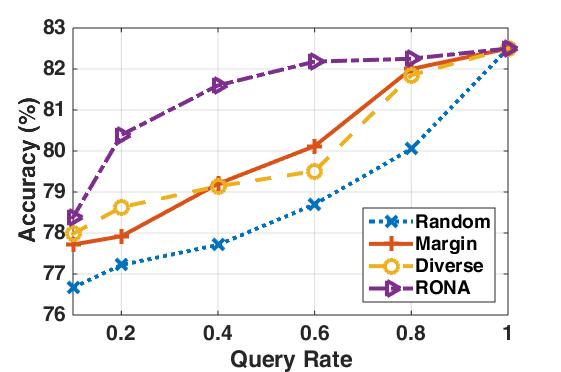}}\quad
\subfloat[Testing Accuracy]{\includegraphics[width=2.15in]{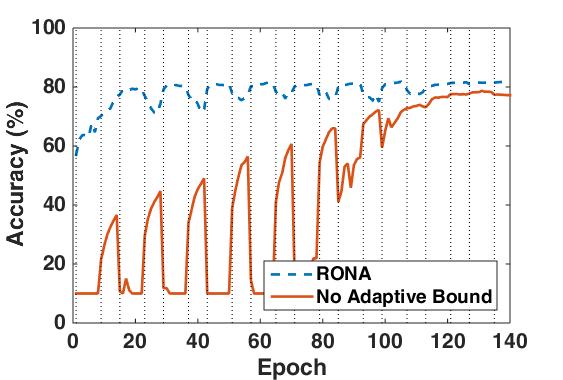}}
\caption{Performance impact of parameters.} \label{fig:para}
\end{figure*}
%======================================================

\section{Experimental Evaluation}
The framework \system\ is evaluated based on three popular image datasets: MNIST \cite{LeCun1998}, SVHN \cite{Netzer2011}, and CIFAR-10 \cite{Krizhevsky2009}. We first use CIFAR-10 to examine the performance impact of different parameters and the effectiveness of proposed techniques in \system. Then we verify privacy protection and compression performance based on MNIST, SVHN, and CIFAR-10. The experiment details are reported in the GitHub repository together with the codes for reproducibility.

\subsection{Effect of Parameters}
We use CIFAR-10 in this group of experiments. CIFAR-10 contains 50K training samples belonging to 10 classes. We randomly choose 80\% training samples as the public data while the reset 20\% as the sensitive data. We preprocess the dataset by normalizing each sample. The widely used convolutional deep neural network, Conv-Large \cite{Laine2017,Park2017}, is pretrained as the cumbersome teacher model on both public data and sensitive data. A modified Conv-Small network is used as the compact student model that will be trained based on \system. The performance of the compact student model is affected by multiple parameters. We examine them individually, keeping the others constant, to show their effects.

\textbf{Hint learning epochs}. It can be observed from Fig. \ref{fig:para}(a) that the accuracy of the student model increases when the hint learning epoch ascends. But the increase diminishes as the hint learning epoch becomes large, especially when $RT_d$, \textit{i.e}., the total epochs of distillation learning, is small. As the hint learning consumes the limited privacy budget, it is not appropriate to set an overly large value of hint learning epoch.

\textbf{Iterations for distillation learning}. The total epochs of distillation learning are determined by the rounds of iterations $R$ and the epochs per iteration $T_d$. Generally, as shown in Fig. \ref{fig:para}(b), a larger value of $RT_d$ brings a more effective student model because the student model can learn more knowledge from the teacher. When $RT_d$ is fixed, $R$ should be set as a moderate value.

\textbf{Batch size}. Fig. \ref{fig:para}(c) shows that the student's performance descends with the increase of batch size. In \system, the student queries the teacher in a batch-by-batch mode. A small value of batch size indicates that the teacher would be queried more times, and thus the privacy loss would be high. To achieve a balance between performance and privacy, we set the batch size as 512 in our experiments.

\textbf{Noise scale}. The student model benefits from the additional noise when the noise scale is moderate as shown in Fig. \ref{fig:para}(d). As a DNN usually suffers from the overfitting problem, the norm bound and additional noise act as regularization roles during training. Even when the noise scale is relatively large ($\sigma=20$), the accuracy degradation is less than 1\%. This property is encouraging as more noise can be injected to provide a stronger privacy guarantee per query.

\textbf{Number of query samples}. The experimental results in Fig. \ref{fig:para}(e) show that the student's performance rises when more query samples are available. Nonetheless, more query samples mean a higher privacy loss. The query sample selection method proposed in our work can achieve decent performance by only using 20\% public samples as the query samples. We compare it with three other selection methods: random, margin \cite{wang2017cost}, and diverse \cite{wang2015querying}. Fig. \ref{fig:para}(e) demonstrates the superiority of our proposed query sample selection method.

%=====================================================
\begin{figure*}[tb]
\centering
\subfloat[MNIST]{\includegraphics[width=2.15in]{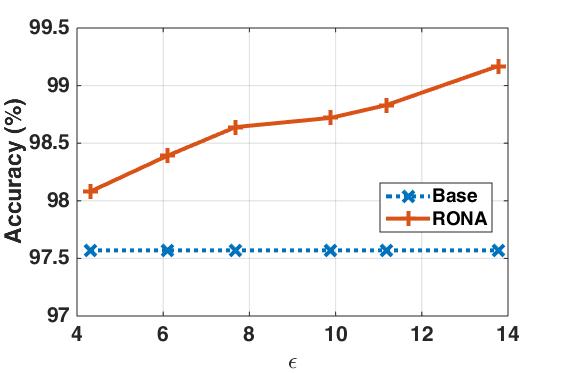}}\quad
\subfloat[SVHN]{\includegraphics[width=2.15in]{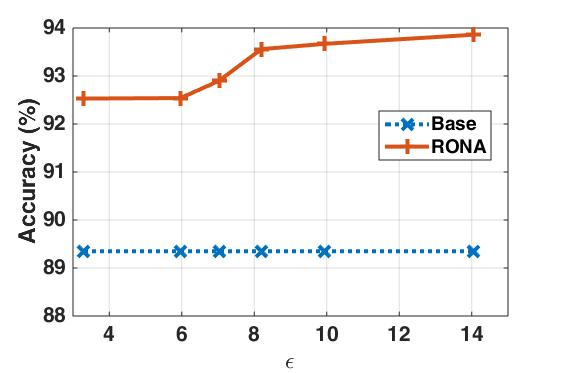}}\quad
\subfloat[CIFAR]{\includegraphics[width=2.19in]{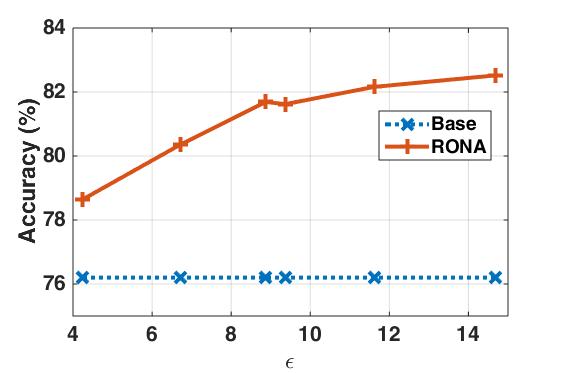}}
\caption{Accuracy vs. privacy budget $\epsilon$. Base denotes the student model trained without querying the teacher model.} \label{fig:budget}
\end{figure*}
%======================================================

\textbf{Adaptive norm bound}. We plot the testing accuracy during the distillation and self learning stage in Fig. \ref{fig:para}(f). Compared with training with the pre-set norm bound, the adaptive norm bound method brings significant performance improvement. It greatly accelerates the training process, obtaining a higher accuracy with much fewer query epochs (\textit{i.e.}, less privacy loss). Besides, we can find that the self learning contributes to the training acceleration as well. At the beginning of the distillation and self learning stage, the accuracy is quickly improved by the self learning without consuming privacy budget.

\subsection{Privacy Protection}

We verify the privacy protection on MNIST, SVHN, and CIFAR-10. MNIST and SVHN are digit image datasets consisting of 60K and 73K training samples, respectively. We randomly choose 40\% SVHN training samples and MNIST training samples as the public data. For MNIST, we use a modified Conv-Small network as the teacher. For SVHN, we use the Conv-Middle network as the teacher.

%======================================================
\begin{table}[tb]
\centering
\caption{Performance of Masking Specific Samples}
 \label{table:mask}
\begin{tabular}{l|c|rrr}
\hline
                                                                                   & \multirow{2}{*}{Base}   & \multicolumn{3}{c}{RONA}                                                                 \\
                                                                                   &                            & \multicolumn{1}{l}{$\epsilon=5.2$} & \multicolumn{1}{l}{$\epsilon=8.7$} & \multicolumn{1}{l}{$\epsilon=29.8$} \\ \hline
\multicolumn{1}{c|}{Overall}                                                       & \multicolumn{1}{r|}{79.87} & 80.93                              & 88.48                              & 90.35                               \\ \hline
\multicolumn{1}{c|}{\begin{tabular}[c]{@{}c@{}}Unseen Classes\end{tabular}} & \multicolumn{1}{r|}{0.00}  & 9.59                               & 46.75                              & 53.43                               \\ \hline
\end{tabular}
\end{table}
%======================================================

\system\ achieves accuracies of 98.64\% and 92.90\% on MNIST and SVHN with $(7.68,10^{-5})$ and $(7.03,10^{-6})$ differential privacy. It outperforms the results in \cite{papernot2017semi} which achieved 98.10\% and 90.66\% accuracy on $(8.03,10^{-5})$ guaranteed MNIST and $(8.19,10^{-6})$ guaranteed SVHN. It is comparable with the latest results in \cite{papernot2018scalable}, achieving 98.5\% and 91.6\% accuracy on $(1.97,10^{-5})$ guaranteed MNIST and $(4.96,10^{-6})$ guaranteed SVHN, which however used better and larger baseline networks. On CIFAR-10, we obtain 81.69\% accuracy with $(8.87,10^{-5})$ privacy, which outperforms 73\% accuracy with $(8.00,10^{-5})$ privacy in \cite{abadi2016deep}. \system's performance is even better than the latest cloud-based solution that achieved 79.52\% accuracy on CIFAR-10 \cite{wang2018not}.

We verify the performance with different privacy budgets on three datasets. Fig. \ref{fig:budget} demonstrates that the accuracies on three datasets generally rise with the increase of privacy budget. Thanks to the techniques proposed in this work, the student can still benefit from the knowledge distillation even when the knowledge is protected by a strong privacy.

In the above experiments, we select the public data randomly from the original training samples. We further test \system\ in a much tougher condition on MNIST where we regard all training samples of the digits 6 and 9 as sensitive data. This data masking mimics a possible contingency in reality when some specific kinds of samples are highly sensitive. For the student model, 6 and 9 are mythical digits it has never seen. As listed in Table \ref{table:mask}, without the teacher's knowledge, the student cannot recognize 6 and 9, getting accuracy of 0. \system\ can significantly improve the student's accuracy on 6 and 9 with a reasonable privacy loss even though the student has never seen 6 and 9 during training.

%======================================================
\begin{table}[tb]
\centering
\caption{Compression Performance} \label{table:compression}
\begin{tabular}{c|c|rrr}
\hline
\multicolumn{1}{l}{}                                                & \multicolumn{1}{l|}{} & \multicolumn{1}{c}{\# Params} & \multicolumn{1}{c}{Time(s)} & \multicolumn{1}{c}{Acc(\%)} \\ \hline
\multirow{3}{*}{MNIST}                                              & T                     & 155.21K                       & 0.76                        & 99.48                       \\
                                                                    & S1                    & 4.97K                         & 0.03                        & 98.94                       \\
                                                                    & S2                    & 9.88K                         & 0.07                        & 99.28                       \\ \hline \hline
\multirow{3}{*}{SVHN}                                               & T                     & 1.41M                         & 7.34                        & 96.36                       \\
                                                                    & S1                    & 0.04M                         & 0.29                        & 94.49                       \\
                                                                    & S2                    & 0.07M                         & 0.39                        & 95.39                       \\ \hline \hline
\multirow{3}{*}{\begin{tabular}[c]{@{}c@{}}CIFAR-10\end{tabular}} & T                     & 3.12M                         & 13.92                       & 86.35                       \\
                                                                    & S1                    & 0.15M                         & 0.93                        & 82.14                       \\
                                                                    & S2                    & 0.52M                         & 3.10                        & 84.57                       \\ \hline
\end{tabular}
\end{table}
%======================================================

\subsection{Compression Performance}
We randomly choose 80\% training samples as the public data to validate the compression performance on the three datasets. For MNIST, SVHN, and CIFAR-10, we enforce $(9.60,10^{-5})$, $(9.83,10^{-6})$, and $(9.59,10^{-5})$ differential privacy, respectively. In order to examine the runtime on mobile devices, we deploy these DNNs on HUAWEI HONOR 8 equipped with ARM Cortex-A53@2.3GHz and Cortex-A53@1.81GHz to process 100 images consecutively.

The results listed in Table \ref{table:compression} show that the models with larger sizes achieve better performance. On all the three datasets, the student models trained by \system\ obtain comparable accuracies to the teacher models in spite of using much less capacity and fewer training data. On MNIST, the student achieves 15$\times$ compression ratio and 11 $\times$ speed-up with only 0.2\% accuracy decrease. On SVHN, the student model obtains slightly worse result ($<1\%$) than the teacher model, while requiring 20 times fewer parameters and 19 times less runtime. On CIFAR-10, the accuracy decreases less than 2\% while the model size is 6 times smaller. The above results argue that \system\ can privately compress large models with acceptable accuracy loss.

\bibliographystyle{aaai}
\bibliography{sample-bibliography}

\begin{thebibliography}{}

\bibitem[\protect\citeauthoryear{Abadi \bgroup et al\mbox.\egroup
  }{2016}]{abadi2016deep}
Abadi, M.; Chu, A.; Goodfellow, I.; McMahan, H.~B.; Mironov, I.; Talwar, K.;
  and Zhang, L.
\newblock 2016.
\newblock Deep learning with differential privacy.
\newblock In {\em Proceedings of the 2016 ACM SIGSAC Conference on Computer and
  Communications Security}, CCS '16,  308--318.

\bibitem[\protect\citeauthoryear{Beimel \bgroup et al\mbox.\egroup
  }{2014}]{beimel2014bounds}
Beimel, A.; Brenner, H.; Kasiviswanathan, S.~P.; and Nissim, K.
\newblock 2014.
\newblock Bounds on the sample complexity for private learning and private data
  release.
\newblock {\em Machine Learning} 94(3):401--437.

\bibitem[\protect\citeauthoryear{Bucilua, Caruana, and
  Niculescu-Mizil}{2006}]{bucila2006model}
Bucilua, C.; Caruana, R.; and Niculescu-Mizil, A.
\newblock 2006.
\newblock Model compression.
\newblock In {\em Proceedings of the 12th ACM SIGKDD International Conference
  on Knowledge Discovery and Data Mining}, KDD '06,  535--541.

\bibitem[\protect\citeauthoryear{Cao \bgroup et al\mbox.\egroup
  }{2017}]{cao2017deepmood}
Cao, B.; Zheng, L.; Zhang, C.; Yu, P.~S.; Piscitello, A.; Zulueta, J.; Ajilore,
  O.; Ryan, K.; and Leow, A.~D.
\newblock 2017.
\newblock Deepmood: modeling mobile phone typing dynamics for mood detection.
\newblock In {\em Proceedings of the 23rd ACM SIGKDD International Conference
  on Knowledge Discovery and Data Mining}, KDD '17,  747--755.

\bibitem[\protect\citeauthoryear{Chen \bgroup et al\mbox.\egroup
  }{2017}]{cheng2018learning}
Chen, G.; Choi, W.; Yu, X.; Han, T.; and Chandraker, M.
\newblock 2017.
\newblock Learning efficient object detection models with knowledge
  distillation.
\newblock In {\em Advances in Neural Information Processing Systems 30}, NIPS
  '17.
\newblock  742--751.

\bibitem[\protect\citeauthoryear{Claerhout and DeMoor}{2005}]{hipaa2013}
Claerhout, B., and DeMoor, G.
\newblock 2005.
\newblock Privacy protection for clinical and genomic data: The use of
  privacy-enhancing techniques in medicine.
\newblock {\em International Journal of Medical Informatics} 74(2):257 -- 265.

\bibitem[\protect\citeauthoryear{Dwork and Roth}{2014}]{dwork2014the}
Dwork, C., and Roth, A.
\newblock 2014.
\newblock The algorithmic foundations of differential privacy.
\newblock {\em Foundations and Trends in Theoretical Computer Science}
  9(3-4):211--407.

\bibitem[\protect\citeauthoryear{Dwork}{2011}]{dwork2011diff}
Dwork, C.
\newblock 2011.
\newblock {\em Differential Privacy}.
\newblock Boston, MA: Springer US.
\newblock  338--340.

\bibitem[\protect\citeauthoryear{Han, Mao, and Dally}{2016}]{han2015deep}
Han, S.; Mao, H.; and Dally, W.~J.
\newblock 2016.
\newblock Deep compression: Compressing deep neural networks with pruning,
  trained quantization and huffman coding.
\newblock In {\em 4th International Conference on Learning Representations},
  ICLR '16.

\bibitem[\protect\citeauthoryear{Hinton, Vinyals, and
  Dean}{2014}]{hinton2015distilling}
Hinton, G.; Vinyals, O.; and Dean, J.
\newblock 2014.
\newblock Distilling the knowledge in a neural network.
\newblock In {\em Advances in Neural Information Processing Systems Deep
  Learning Workshop}, NIPS '14.

\bibitem[\protect\citeauthoryear{Hitaj, Ateniese, and
  Perez-Cruz}{2017}]{hitaj2017deep}
Hitaj, B.; Ateniese, G.; and Perez-Cruz, F.
\newblock 2017.
\newblock Deep models under the gan: Information leakage from collaborative
  deep learning.
\newblock In {\em Proceedings of the 2017 ACM SIGSAC Conference on Computer and
  Communications Security}, CCS '17,  603--618.

\bibitem[\protect\citeauthoryear{Howard \bgroup et al\mbox.\egroup
  }{2017}]{howard2017mobile}
Howard, A.~G.; Zhu, M.; Chen, B.; Kalenichenko, D.; Wang, W.; Weyand, T.;
  Andreetto, M.; and Adam, H.
\newblock 2017.
\newblock Mobilenets: Efficient convolutional neural networks for mobile vision
  applications.
\newblock {\em arXiv:1704.04861}.

\bibitem[\protect\citeauthoryear{Korupolu, Plaxton, and
  Rajaraman}{1998}]{korupolu2000analysis}
Korupolu, M.~R.; Plaxton, C.~G.; and Rajaraman, R.
\newblock 1998.
\newblock Analysis of a local search heuristic for facility location problems.
\newblock In {\em Proceedings of the Ninth Annual ACM-SIAM Symposium on
  Discrete Algorithms}, SODA '98,  1--10.

\bibitem[\protect\citeauthoryear{Krizhevsky and Hinton}{2009}]{Krizhevsky2009}
Krizhevsky, A., and Hinton, G.
\newblock 2009.
\newblock Learning multiple layers of features from tiny images.

\bibitem[\protect\citeauthoryear{Laine}{2017}]{Laine2017}
Laine, S.
\newblock 2017.
\newblock Temporal ensembling for semi-supervised learning.
\newblock In {\em 5th International Conference on Learning Representations},
  ICLR '17.

\bibitem[\protect\citeauthoryear{LeCun \bgroup et al\mbox.\egroup
  }{1998}]{LeCun1998}
LeCun, Y.; Bottou, L.; Bengio, Y.; and Hinton, G.
\newblock 1998.
\newblock Gradient-based learning applied to document recognition.
\newblock {\em Proceedings of the IEEE} 86(11):2278--2324.

\bibitem[\protect\citeauthoryear{Lee}{2017}]{lee2017technology}
Lee, P.
\newblock 2017.
\newblock Technology, media and telecommunications predictions.
\newblock {\em Delloitte Touche Tohmatsu Limited}.

\bibitem[\protect\citeauthoryear{Li \bgroup et al\mbox.\egroup
  }{2016}]{li2016droidclassifier}
Li, Z.; Sun, L.; Yan, Q.; Srisa-an, W.; and Chen, Z.
\newblock 2016.
\newblock Droidclassifier: Efficient adaptive mining of application-layer
  header for classifying android malware.
\newblock In {\em International Conference on Security and Privacy in
  Communication Systems},  597--616.

\bibitem[\protect\citeauthoryear{Netzer \bgroup et al\mbox.\egroup
  }{2011}]{Netzer2011}
Netzer, Y.; Wang, T.; Coates, A.; Bissacco, A.; and Ng, B. W. A.~Y.
\newblock 2011.
\newblock Reading digits in natural images with unsupervised feature learning.
\newblock In {\em Proceedings of the 24th International Conference on Neural
  Information Processing Systems}, NIPS '11.
\newblock  1--9.

\bibitem[\protect\citeauthoryear{Osia \bgroup et al\mbox.\egroup
  }{2017}]{osia2017a}
Osia, S.~A.; Shamsabadi, A.~S.; Taheri, A.; Rabiee, H.~R.; Lane, N.~D.; and
  Haddadi, H.
\newblock 2017.
\newblock A hybrid deep learning architecture for privacy-preserving mobile
  analytics.
\newblock {\em arXiv:1703.02952}.

\bibitem[\protect\citeauthoryear{Papernot \bgroup et al\mbox.\egroup
  }{2017}]{papernot2017semi}
Papernot, N.; Abadi, M.; Erlingsson, U.; Goodfellow, I.; and Talwar, K.
\newblock 2017.
\newblock Semi-supervised knowledge transfer for deep learning from private
  training data.
\newblock In {\em 5th International Conference on Learning Representations},
  ICLR '17.

\bibitem[\protect\citeauthoryear{Papernot \bgroup et al\mbox.\egroup
  }{2018}]{papernot2018scalable}
Papernot, N.; Song, S.; Mironov, I.; Raghunathan, A.; Talwar, K.; and
  Erlingsson, U.
\newblock 2018.
\newblock Scalable private learning with pate.
\newblock In {\em 6th International Conference on Learning Representations},
  ICLR '18.

\bibitem[\protect\citeauthoryear{Park \bgroup et al\mbox.\egroup
  }{2017}]{Park2017}
Park, S.; Park, J.-K.; Shin, S.-J.; and Moon, I.-C.
\newblock 2017.
\newblock Adversarial dropout for supervised and semi-supervised learning.
\newblock {\em arXiv:1707.03631}.

\bibitem[\protect\citeauthoryear{Romero \bgroup et al\mbox.\egroup
  }{2015}]{romero2015fitnets}
Romero, A.; Ballas, N.; Kahou, S.~E.; Chassang, A.; Gatta, C.; and Bengio, Y.
\newblock 2015.
\newblock Fitnets: Hints for thin deep nets.
\newblock In {\em 3th International Conference on Learning Representations},
  ICLR '15.

\bibitem[\protect\citeauthoryear{Triastcyn and
  Faltings}{2018}]{triastcyn2018generating}
Triastcyn, A., and Faltings, B.
\newblock 2018.
\newblock Generating artificial data for private deep learning.
\newblock {\em arXiv:1803.03148}.

\bibitem[\protect\citeauthoryear{Wang and Ye}{2013}]{wang2015querying}
Wang, Z., and Ye, J.
\newblock 2013.
\newblock Querying discriminative and representative samples for batch mode
  active learning.
\newblock In {\em Proceedings of the 19th ACM SIGKDD International Conference
  on Knowledge Discovery and Data Mining}, KDD '13,  158--166.

\bibitem[\protect\citeauthoryear{Wang \bgroup et al\mbox.\egroup
  }{2017}]{wang2017cost}
Wang, K.; Zhang, D.; Li, Y.; Zhang, R.; and Lin, L.
\newblock 2017.
\newblock Cost-effective active learning for deep image classification.
\newblock {\em IEEE Transactions on Circuits and Systems for Video Technology}
  27(12):2591--2600.

\bibitem[\protect\citeauthoryear{Wang \bgroup et al\mbox.\egroup
  }{2018}]{wang2018not}
Wang, J.; Zhang, J.; Bao, W.; Zhu, X.; Cao, B.; and Yu, P.~S.
\newblock 2018.
\newblock Not just privacy: Improving performance of private deep learning in
  mobile cloud.
\newblock In {\em Proceedings of the 24th ACM SIGKDD International Conference
  on Knowledge Discovery \& Data Mining}, KDD '18,  2407--2416.

\end{thebibliography}

\end{document}